\documentclass{article}
\usepackage{spconf,amsmath,epsfig}
\usepackage{slashbox}
\usepackage{cite}
\usepackage[]{hyperref}
\usepackage{microtype}
\newcommand{\etal}{\textit{et al}. }

\usepackage{color}
\usepackage{multirow}

\let\OLDthebibliography\thebibliography
\renewcommand\thebibliography[1]{
  \OLDthebibliography{#1}
  \setlength{\parskip}{0pt}
  \setlength{\itemsep}{0pt plus 0.3ex}
}

\begin{document}\sloppy

% Example definitions.
% --------------------
\def\x{{\mathbf x}}
\def\L{{\cal L}}

% Title.
% ------
\title{DENSE POINT PREDICTION: A SIMPLE BASELINE FOR CROWD \\ COUNTING AND LOCALIZATION}
%
% Single address.
% ---------------
\name{Yi~Wang, Xinyu~Hou, and Lap-Pui~Chau}
%Address and e-mail should NOT be added in the submission paper. They should be present only in the camera ready paper. 
\address{School of Electrical and Electronics Engineering, Nanyang Technological University, Singapore \\
E-mail: wang1241@e.ntu.edu.sg, houx0008@e.ntu.edu.sg, elpchau@ntu.edu.sg}

\maketitle

\begin{abstract}
In this paper, we propose a simple yet effective crowd counting and localization network named SCALNet. Unlike most existing works that separate the counting and localization tasks, we consider those tasks as a pixel-wise dense prediction problem and integrate them into an end-to-end framework. Specifically, for crowd counting, we adopt a counting head supervised by the Mean Square Error (MSE) loss. For crowd localization, the key insight is to recognize the keypoint of people, i.e., the center point of heads. We propose a localization head to distinguish dense crowds trained by two loss functions, i.e., Negative-Suppressed Focal (NSF) loss and False-Positive (FP) loss, which balances the positive/negative examples and handles the false-positive predictions. Experiments on the recent and large-scale benchmark, NWPU-Crowd, show that our approach outperforms the state-of-the-art methods by more than 5\% and 10\% improvement in crowd localization and counting tasks, respectively. The code is publicly available at \href{https://github.com/WangyiNTU/SCALNet}{https://github.com/WangyiNTU/SCALNet}.
\end{abstract}
\section{Introduction}
\label{sec:intro}

\begin{figure} [t]
\centering 
\includegraphics[width=8.5cm]{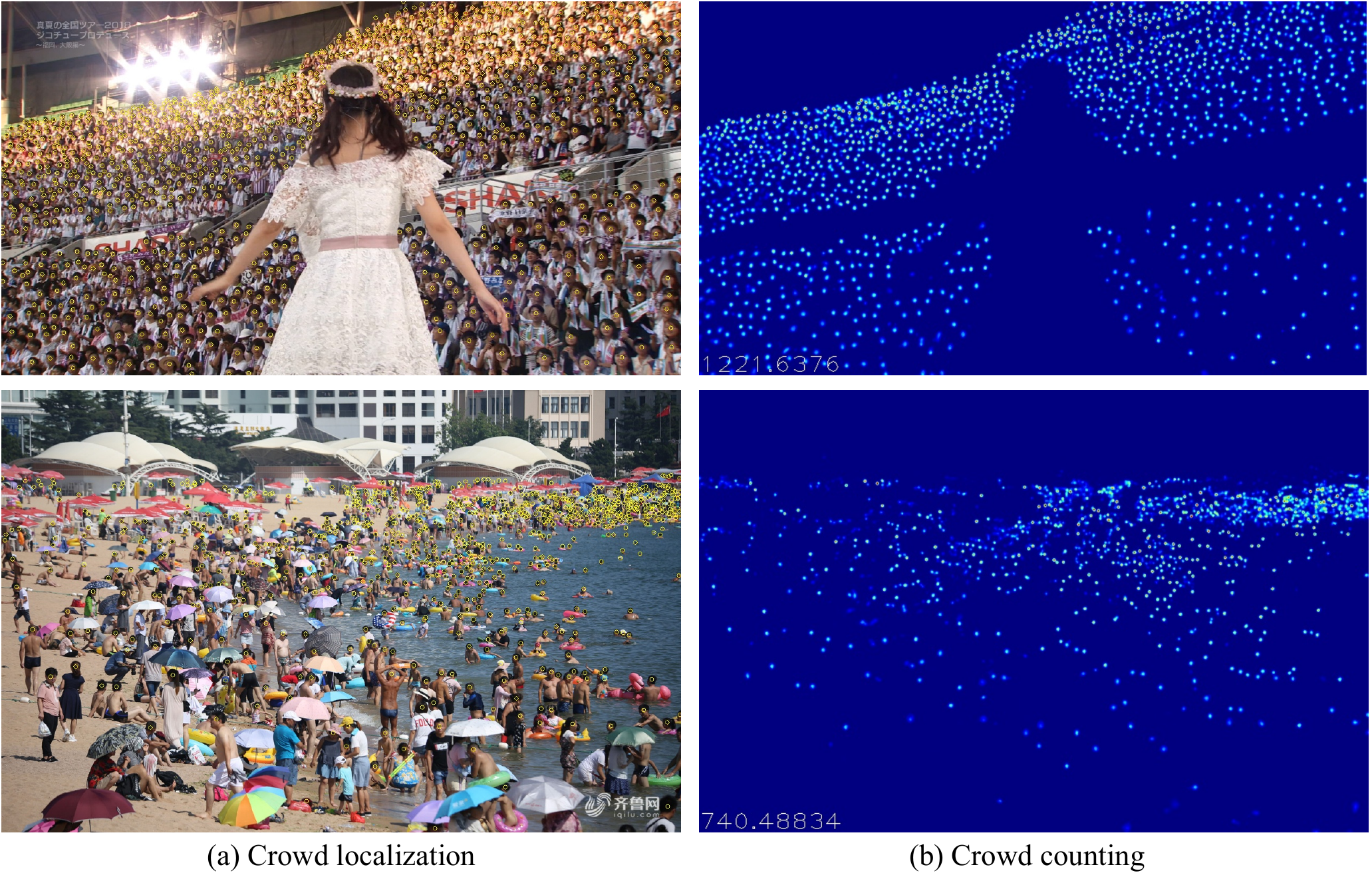}
\caption{Illustration of crowd counting and localization. (a) Crowd localization (yellow circles) by the localization head of the proposed method. (b) Crowd counting by the counting head of the proposed method, where the predicted counts (1222 and 740) are shown at the bottom-left corner for the top and bottom images, respectively. The ground-truth counts are 1186 and 773. \textit{Zoom in the figure for better viewing.}}
\label{fig1}
\end{figure}

Crowd counting and crowd localization tasks, with a wide range of real-life applications, have attracted much attention in recent years. The count and location information are crucial for analyzing crowded pedestrians, traffic flow, and cells \cite{ccsurvey}. Many researchers applied Deep Neural Networks (DNNs) to achieve high performance in crowd counting. Recently, some researchers commence locating crowded objects beyond counting \cite{RAZ_Loc}. Examples of crowd counting and localization in dense crowds are shown in Fig. \ref{fig1}. 

Traditionally, crowd counting and localization are considered as two separate tasks. State-of-the-art crowd counting methods are based on regression networks which focus on regressing the counts but ignore the locations of object instances. These methods aim to predict a density map wherein the count is calculated by the integral over the map \cite{MCNN,DSSINet,wang2019object}, as shown in Fig. \ref{fig1} (b). Congested Scene Recognition Network (CSRNet) \cite{CSRNet} utilizes VGG-16 \cite{simonyan2014very} as the frontend, followed by the dilated convolution layers as the backend to generate a density map. Reference \cite{SCAR} emphasized the contextual information of the whole image by adding a spatial-/channel-wise attention regression (SCAR) to CNN. Chen \etal \cite{SRFNet} proposed a two-stage architecture consisting of a band-pass stage and a rolling guidance stage to better extract multi-scale features with multi-level information. Other regression-based works (like \cite{SANet,CANNet,BL,SFCN}) and detailed comparisons can refer to the survey paper \cite{ccsurvey} and benchmark \cite{nwpucrowd}.

Owing to the highly-occluded crowds and point-level annotations (heads are labeled by points), traditional object detection (\cite{fasterrcnn,tinyface}) methods produce low recall of crowds, damaging the counting performance. Recently, crowd localization methods have been developed for dense crowds, like \cite{RAZ_Loc,blob,PIBO}, showing that those methods can localize crowds as well as obtain a comparable counting result. The localization process is to produce the keypoints \cite{RAZ_Loc,dotprediction}, blobs \cite{blob}, or bounding boxes \cite{PIBO,LSC,wang2020selftraining} on each head (see Fig. \ref{fig1} (a)), and the total count is obtained by a confidence thresholding.

\begin{table*}[t]
\begin{center}
\caption{Two basic crowd tasks addressed by dense regression-based methods and dense classification-based methods.} \label{table:comp}
\begin{tabular}{|l|c|c|}
\hline
 \backslashbox[66mm]{Method}{Task}                    & Crowd counting                           & Crowd localization                 \\ \hline
Density map prediction \cite{ccsurvey,MCNN,wang2019object,CSRNet,SCAR,SRFNet,SANet,CANNet,BL,SFCN}   & Dense regression & Post-processing of density map \cite{compositionloss, domainadaptive}      \\ \hline
Keypoint, blob, or box prediction \cite{RAZ_Loc,dotprediction,blob,PIBO,LSC,wang2020selftraining} & Confidence thresholding & Dense classification \\ \hline
\end{tabular}
\vspace{-4mm}
\end{center}
\end{table*}

However, crowd counting and localization are relatively independent in previous works. Each category has a specific network architecture and training strategy. The main diagonal of Table \ref{table:comp} shows that crowd counting is addressed by regression-based density map prediction, while crowd localization is resolved by classification-based keypoint, blob, or box prediction. To achieve the other task, post-processing techniques (like local peak selection \cite{compositionloss} and Gaussian-prior Reconstruction, GPR \cite{domainadaptive}) performed on top of the generated density map help regression methods accomplish the crowd localization task, while the confidence thresholding enables classification methods to count objects, as shown in the counter diagonal of Table \ref{table:comp}. These post-processing techniques are performed separately from the neural networks, lacking an end-to-end training and inference.

\begin{figure} [t]
\centering 
\includegraphics[width=8.2cm]{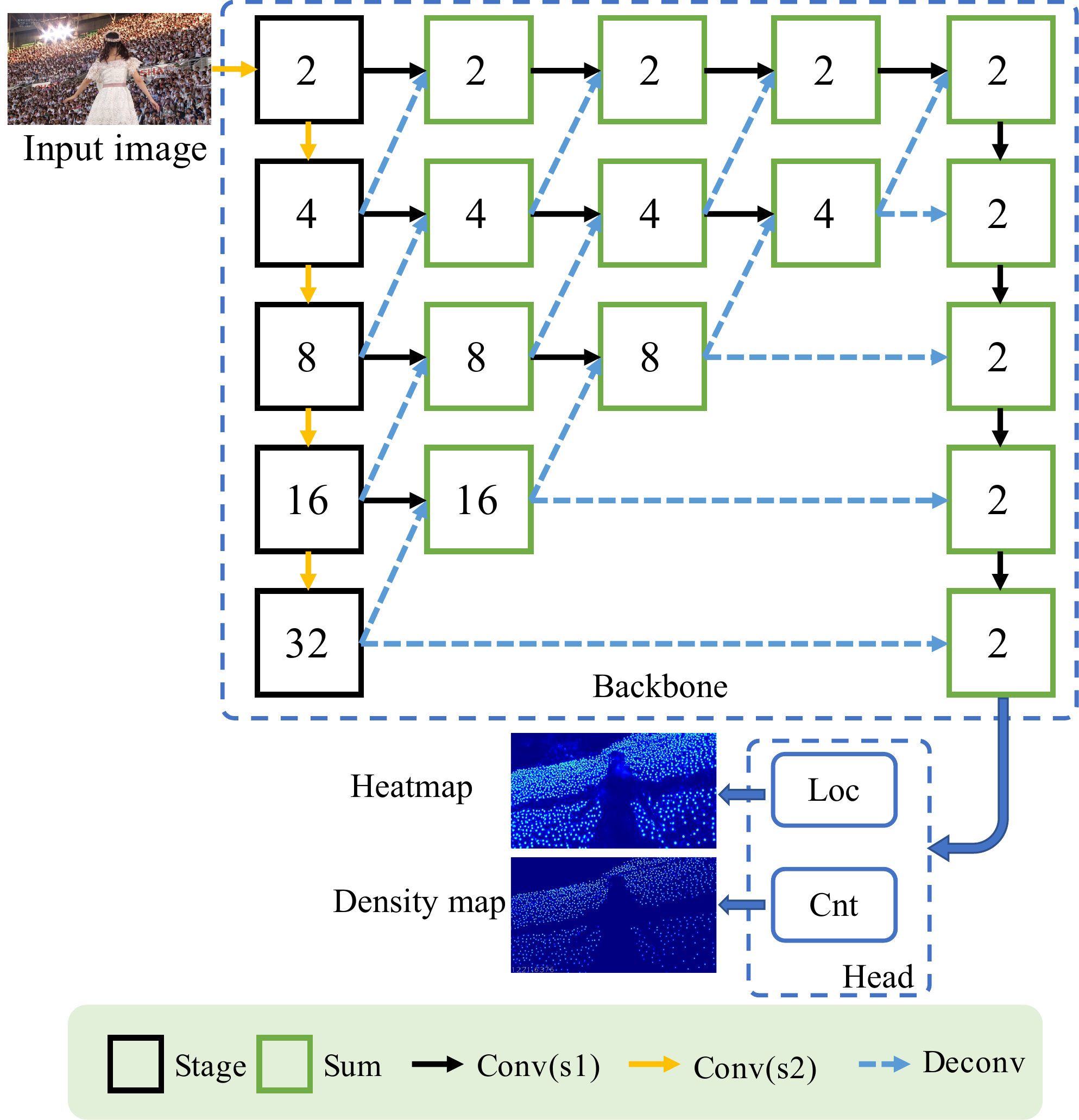}
\caption{Network architecture of SCALNet. The numbers inside the nodes denote the stride value, indicating the feature size compared with the input size (e.g., the stride of 2 means 1/2 of input size). Sum nodes indicate pixel-by-pixel summation of two features. Conv(s$x$) denotes the convolution with the stride of $x$. Deconv means the transposed convolution (or deconvolution).}
\label{fig:network}
\end{figure}

From the perspective of dense prediction \cite{DensePrediction}, crowd counting and localization are relevant since they either predict continuous or discrete labels for each pixel of an image. Therefore, we consider both tasks as a dense prediction problem and build an end-to-end CNN to predict two dense feature maps. One contains the continuous labels (i.e., the density map), and the other contains the discrete labels (i.e., the heatmap with pixel-wise classification). Specifically, we propose a Simple yet effective Counting And Localization Network, named SCALNet, as shown in Fig. \ref{fig:network}. For crowd counting, we build a simple counting head network supervised by the Mean Square Error (MSE) loss for density map regression. For crowd localization, we propose a keypoint localization network to recognize each human head of crowds. The supervisory information is the point-level annotations. To stably train the network, we propose to use two classification losses, i.e., Negative-Suppressed Focal (NSF) loss and False-Positive (FP) loss, which can balance the positive/negative examples and handle the false-positive predictions. The human heads' locations can be obtained by decoding the local peaks on the output heatmap. 

We evaluate our approach on the recent large-scale benchmark, NWPU-Crowd \cite{nwpucrowd}. Experimental results show that our approach outperforms state-of-the-art methods by more than 5\% improvement in crowd localization and 18\% (resp. 12\%) improvement on MAE (resp. MSE) in crowd counting.

The remainder of this paper is organized as follows. In Section \ref{sec:method}, we present our approach in detail. Then, we conduct experiments and discuss the effectiveness of our approach in Section \ref{sec:results}. Finally, we conclude our paper in Section \ref{sec:conclusion}.

\section{Proposed Method}
\label{sec:method}

In this section, we present the proposed method, including network architecture, supervision, and loss functions. 

\subsection{Network Architecture}
\label{subsec:network}
We consider the crowd localization problem as a dense classification (i.e., pixel-wise binary classification) and the crowd counting problem as a dense regression, such that many dense prediction networks can be employed as the backbone. Our network architecture is shown in Fig. \ref{fig:network}.

\textbf{Backbone Network.} Based on CenterNet \cite{OasP}, we modify the Deep Layer Aggregation (DLA-34) \cite{DLA} structure as the backbone network. The backbone provides a stage-wise fusion manner for multi-scale features from the stride of 32 to 2. Small-scale features are gradually fused with 2$\times$up large-scale features, leading to the fusion of semantic information and spatial information. Instead of the original 4-stride DLA-34, we utilize the 2-stride features in the fusion step to retain more spatial information, as shown at the top row of the network in Fig. \ref{fig:network}. Finally, different-level features are aggregated into 2-stride features by the deconvolutional layers and convolutional layers, as shown at the right column of the network. This modification significantly improves the localization ability with a little inference time increase (see Section \ref{subsec:abl}).

\textbf{Localization Head Network (LHN).} Since crowds are densely distributed, the distance between two heads may be less than several pixels in an image. To distinguish those heads, LHN should output the full-resolution heatmap. Specifically, the heatmap size is set to be the same as the input size. Connected on the top layer of the backbone network, LHN is the sequential layers of Dconv(k4-s2)-Conv(k3-s1)-BN-ReLU-Conv(k1-s1)-Sigmoid, where Dconv(k4-s2) is a deconvolutional layer with the kernel of 4 and stride of 2, Conv is a convolutional layer, BN is the batch normalization, and ReLU (Rectified Linear Unit) and Sigmoid are the activation functions.

\textbf{Counting Head Network (CHN).} CHN aims to produce a density map by regression. The density map can accumulate the density values of crowded objects, so the low-resolution map satisfies the counting task. Specifically, CHN is fed by the output features of the backbone network and generates a 2-stride density map. The structure of CHN consists of Conv(k3-s1)-PReLU-Conv(k1-s1)-PReLU, where PReLU stands for the parametric ReLU. 

\textbf{Inference.} Throughout a forward pass of the network, the output heatmap contains the human head's estimate. The heads exist in the high-confidence points (peaks) of the heatmap. We can see that the heatmap in Fig. \ref{fig:network} contains sharp peaks, demonstrating a good localization ability. Following \cite{OasP}, we employ a 3$\times$3 local maximum operation to filter out noisy points around the peaks and select the peaks by a confidence thresholding as heads' estimate. The threshold is obtained by searching a confidence interval \([0.3, 0.5]\) where the selected threshold achieves the best localization performance on the validation set. The heads' locations are obtained by the coordinates of the peaks. From the counting head network, the headcount is obtained by simply summing all density values over the output density map.

\subsection{Supervision}
\textbf{Localization supervision.} We first generate the ground-truth dot map from the point-level annotations of the benchmark. The point annotations provide the coordinates of heads in an image. To be specific, we initialize each pixel value to 0 on the dot map, representing the negative labels. We set the pixel values to 1 according to the heads' coordinates, representing the positive labels. Then, we follows \cite{OasP} to generate the heatmap as localization supervision. The Gaussian kernels are placed at the head locations in order to smooth the supervision around the positive labels. We select the maximum value when two Gaussian kernels overlap. The Gaussian kernel is defined by
\begin{equation}
K^{l}_{xy}=\exp \left (-\frac{(x-c_{x})^{2}+(y-c_{y})^2}{2\sigma_{d}^{2}} \right ),
\end{equation}
where $(x,y)$ is the pixel coordinate, $(c_{x},c_{y})$ is the coordinate of a head, and $\sigma_{d}$ is the scale-related standard deviation calculated by its distance to the $3$ nearest heads \cite{MCNN}, i.e., $\sigma_{d}=0.1\cdot \sum_{i=1}^{3}d_{i}$, where $d_{i}$ is the distance to the \(i\)-th head.

\textbf{Counting supervision.} We generate the ground-truth density map as counting supervision. A normalized Gaussian kernel with the integration of 1 is accumulated at a head location. The normalized Gaussian kernel is defined by
\begin{equation}
K^{c}_{xy}=\frac{1}{\sqrt{2\pi }\sigma_{c}}\exp \left (-\frac{(x-c_{x})^{2}+(y-c_{y})^2}{2\sigma_{c}^{2}} \right ),
\end{equation}
where $\sigma_{c}$ is a constant standard deviation. We empirically set $\sigma_{c}=3$ for all experiments, which makes the kernel sharp.

\subsection{Loss Functions}
\textbf{Localization loss.} We employ a pixel-wise classification loss. The loss needs to handle the imbalance of training examples and the complex background. First, the number of positive examples (heads) is much less than that of negative examples (background). For the case of 1,000 pedestrians in a 1080$\times$1080 image, the ratio of positive/negative is about 1:1000. Such an imbalance brings a challenge to the training process. Second, the complex background may cause many false-positive head detections.

In view of these issues, we propose to utilize two complementary loss functions, i.e., Negative-Suppressed Focal (NSF) loss and False-Positive (FP) loss, to stabilize the training process. The NSF loss adopts the focal cross-entropy loss with negative example suppression, like \cite{wang2020selftraining}: 
\begin{equation}\label{eq:nsf}
L_{nsf}= -\frac{1}{M}\sum_{j\in I}\begin{cases}
(1-\hat{p}_{j})^{\gamma }log(\hat{p}_{j}), \qquad \qquad \quad \text{ if } p=1, \\ 
\frac{1}{16} (1-p_{j})^{\delta}(\hat{p}_{j})^{\gamma }log(1-\hat{p}_{j}), \textrm{otherwise}, 
\end{cases}
\end{equation}
where $M$ is the number of positive examples in image $I$, \(\hat{p}_{j}\) is the predicted probability, \(p_{j}\) is the ground truth with 1 for the head and 0 for the background, \(\gamma\) is the parameter of the focal loss \cite{focalloss} to control the weights of well-classified examples, and \(\delta\) is the parameter to decrease the punishments of the negative labels around the positive ones. $\frac{1}{16}$ is the down-weight to suppress the contribution of negative examples. 

However, the NFS loss is a holistic suppression of the negative examples, but the false-positive peaks still appear in the heatmap due to the complex background. Thus, we propose an FP loss to emphasize false-positive detections. Specifically, we find the false-positive region according to the ground-truth heatmap and the predicted heatmap, denoted by $H$ and $\hat{H}$, respectively. The background region is defined by $B=\{(x,y)| H_{xy} = 0 \}$, and the positive region of $\hat{H}$ is $P=\{(x,y)|\hat{H}_{xy} >0.1\}$. Hence, the false-positive region $F$ is formulated by:
\begin{equation}
F=B\cap P.
\end{equation}
Formally, the FP loss is employed in the false-positive region:
\begin{equation}\label{eq:fp}
L_{fp} = -\frac{1}{|F|}\sum_{j\in F}(\hat{p}_{j})^{\gamma }log(1-\hat{p}_{j}),
\end{equation}
where $|F|$ denotes the total number of pixels in $F$, and \(\hat{p}_{j}\) stands for the predicted probability of the $j$-th pixel of $F$. The FP loss encourages the network to reduce the false-positive detections.

\textbf{Regression loss.} We adopt the MSE loss for density map regression. Hence, the regression loss is given as follow:
\begin{equation}
L_{r} = \frac{1}{N}\sum_{j\in N}\left \| D_{j}(I_{i})-D_{j}^{gt} \right \|_{2}^{2}
\end{equation}
where $N$ is the number of pixels in the density map, $I_{i}$ is the $i$-th image, and $D$ and $D^{gt}$ are the predicted and ground-truth density map, respectively.

The total loss is the sum of the NFS, FP, regression losses: 
\begin{equation}\label{eq:totalloss}
L=L_{nfs}+\lambda_{1} L_{fp}+\lambda_{2} L_{r},
\end{equation}
where $\lambda_{1}$ and $\lambda_{2}$ are the hype-parameters to control the weight between those losses.

\section{Experimental Results}
\label{sec:results}

In this section, we first describe the experimental configurations. Then, we compare our approach with state-of-the-art localization and counting methods. Finally, we discuss the effectiveness of the proposed modules.

\subsection{Experimental Configurations}
\textbf{Dataset.} We evaluate our method on the NWPU-Crowd \cite{nwpucrowd} benchmark. The benchmark contains 5109 images and more than 2 million heads with the point and box labels. It is split into the training, validation, and test sets with 3109, 500, and 1500 images, respectively. We train our method in the training set and choose the best model by the validation set. The results in the testing set are obtained by submitting the predictions to the benchmark website. 

\textbf{Metrics.} We employ the evaluation protocol \cite{nwpucrowd} of NWPU-Crowd. For crowd localization, the Precision (Pre.), Recall (Rec.), and F1-measure (F1-m.) are calculated. The higher the value is, the better the performance achieves. The predicted point (the center point of a head) is true positive when its distance to the ground-truth point is less than a predefined distance threshold. Two thresholds are defined by $\sigma_{s}= \min(h,w)/2$ and $\sigma_{l}= \sqrt{h^{2}+w^{2}}/2$, where $h$ and $w$ are the height and width of the head, respectively. The former is harder to reach than the latter. For crowd counting, we used the Mean Absolute Error (MAE), root Mean Squared Error (MSE), and mean Normalized Absolute Error (NAE). The lower the value is, the better the performance achieves.

\textbf{Implementation details.} The network was initialized by the pre-trained weights of DLA-34 \cite{DLA} on ImageNet. We trained the network with the Adam optimizer for 180 epochs. The learning rate was set to 1e-4 and then decayed by 10 at the 135th epoch. The batch size was 32. In Eqs. (\ref{eq:nsf}) and (\ref{eq:fp}), $\gamma$ and $\delta$ were set to 2 and 4, respectively. In Eq. (\ref{eq:totalloss}), $\lambda_{1}$ and $\lambda_{2}$ were empirically set to 1 and 1000, respectively. We down-scale the longer side of images to 1600 pixels if it is larger than 1600. The training examples were augmented by random horizontal flipping, random cropping (320$\times$320), and normalization. The training and testing run on a single RTX 3080 Ti GPU.

\begin{figure*} [t]
\centering 
\includegraphics[width=18cm]{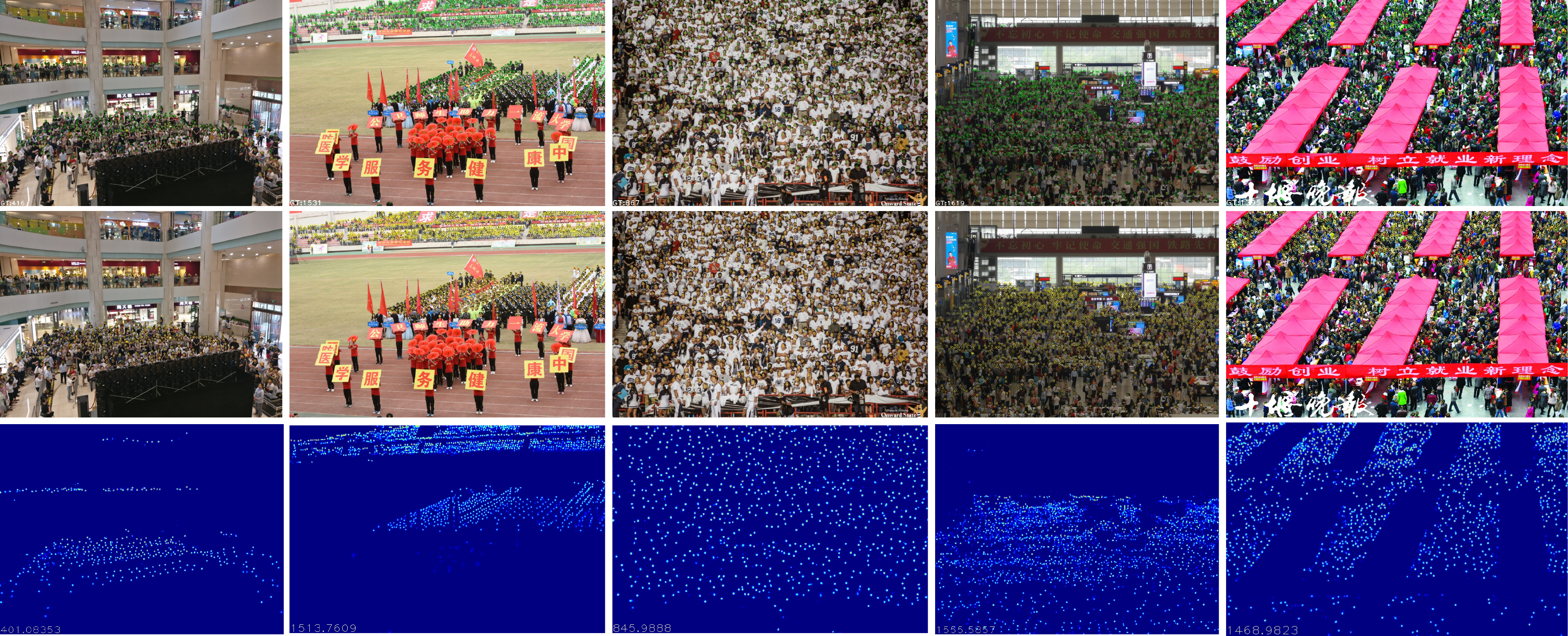}
\caption{Qualitative results of our method. The first row shows ground-truth point annotations (green circles) with the counts shown at the bottom-left corner. The second row shows the predicted head locations (yellow circles). The third row shows the predicted density maps and counts. \textit{Zoom in the figure for better viewing.}}
\label{fig:qualitative}
\end{figure*}

\subsection{Comparison with State-of-the-art Methods}
\textbf{Crowd localization.} We first evaluate our SCALNet in crowd localization. Table \ref{tab:localization} shows the comparison with the state-of-the-art crowd localization methods. With point training labels, the proposed approach achieves the best performance in F1-measure and Recall. The F1-measure and Recall are improved by more than 5\% when compared with the recent method, Crowd-SDNet \cite{wang2020selftraining}. Although Faster RCNN \cite{fasterrcnn} obtains the best Precision, it has the extremely low F1-measure and Recall, which means Faster RCNN precisely detects large-scale persons and fails in most dense crowds (see \cite{nwpucrowd}). Some qualitative results of our method are shown in the second row of Fig. \ref{fig:qualitative}, illustrating that our method can predict the center points of heads accurately in crowded scenes. 

\begin{table}[t]
\centering
\caption{Comparison with the state-of-the-art crowd localization algorithms on the NWPU-Crowd test set. The results are calculated under threshold $\sigma_l$. ``*'' means the results are provided by NWPU-Crowd.}
\label{tab:localization}
\resizebox{\columnwidth}{!}{%
\begin{tabular}{l|c|ccc}
\hline
Method      & Training Labels & F1-m. & Pre. & Rec.\\ \hline
Faster RCNN* \cite{fasterrcnn} & Box             & 6.7       & \textbf{95.8}     & 3.5      \\ 
TinyFaces* \cite{tinyface}   & Box             & 56.7      & 52.9     & 61.1     \\  \hline
VGG+GPR* \cite{domainadaptive}     & Point           & 52.5      & 55.8     & 49.6     \\ 
RAZ\_Loc* \cite{RAZ_Loc}     & Point           & 59.8      & 66.6     & 54.3     \\ 
Crowd-SDNet \cite{wang2020selftraining}     & Point          & 63.7      & 65.1     & 62.4     \\ 
SCALNet (ours)        & Point           &    \textbf{69.1}       &  69.2        &   \textbf{69.0}       \\ \hline
\end{tabular}
}
\end{table}

\textbf{Crowd counting.} Table \ref{tab:counting} shows the counting performance. It can be seen that our method outperforms the state-of-the-art methods in both MAE and MSE by a significant margin, with 18\% and 12\% relative improvement when compared with the second-best BL \cite{BL} and CANNet \cite{CANNet}, respectively. The NAE score of our method is comparable to BL \cite{BL}. Note that our method only employs a simple backbone with counting and localization heads. The third row of Fig. \ref{fig:qualitative} shows the density maps and headcounts predicted by our approach.

\begin{table}[t]
\centering
\caption{Comparison with the state-of-the-art crowd counting methods on the NWPU-Crowd test set. ``*'' means the results are reported by NWPU-Crowd.}
\label{tab:counting}
\begin{tabular}{l|ccc}
\hline
Method       & MAE   & MSE    & NAE   \\ \hline
MCNN* \cite{MCNN}        & 232.5 & 714.6  & 1.063 \\ 
SANet* \cite{SANet}        & 190.6 & 491.4  & 0.991 \\  
CSRNet* \cite{CSRNet}        & 121.3 & 387.8  & 0.604 \\ 
CANNet* \cite{CANNet}        & 106.3 & 386.5  & 0.295 \\ 
SCAR* \cite{SCAR}         & 110.0 & 495.3  & 0.288 \\ 
BL* \cite{BL}           & 105.4 & 454.2  & \textbf{0.203} \\ 
SFCN* \cite{SFCN}        & 105.7 & 424.1  & 0.254 \\ 
SCALNet (ours)        &  \textbf{86.8}     &  \textbf{339.9}      &  0.218     \\ \hline
\end{tabular}
\end{table}

\subsection{Ablation study}
\label{subsec:abl}
We conducted the ablation studies to evaluate the effectiveness of the proposed modules, including the backbone with the stride of 2, and the counting and localization head networks (CHN and LHN). Due to the limitation of submissions on the NWPU-Crowd test set, we present the counting and localization results on the validation set, as shown in Table \ref{tab:abl}. 

\begin{table}[t]
\centering
\caption{Ablation studies on the NWPU-Crowd validation set.}
\label{tab:abl}
\begin{tabular}{l|c|c} 
\hline
Method                                                                      & F1-m./Pre./Rec. (\%)          & MAE/MSE/NAE                          \\ 
\hline
\multirow{2}{*}{\begin{tabular}[c]{@{}l@{}}Backbone w/ \\Stride of 4 \end{tabular}}                                                   & $\sigma_l$: 68.9/71.1/66.8  & \multirow{2}{*}{64.9/231.0/0.227}  \\ 
\cline{2-2}
                                                                            & $\sigma_s$: 62.3/60.5/64.3  &                                                                    \\ 
\hline
\multirow{2}{*}{CHN only} & -              & \multirow{2}{*}{81.8/283.6/0.379}                                                  \\ 
\cline{2-2}
                                                                            & -             &                                                                    \\ 
\hline
\multirow{2}{*}{LHN only} & $\sigma_l$: 72.5/73.1/72.1              & \multirow{2}{*}{-}                                                  \\ 
\cline{2-2}
                                                                            & $\sigma_s$: 67.2/67.7/66.8             &                                                                    \\ 
\hline
\multirow{2}{*}{\begin{tabular}[c]{@{}l@{}}Both heads \\Stride of 2 \end{tabular}}                                                       & $\sigma_l$: 72.4/73.5/71.4             & \multirow{2}{*}{64.4/251.1/0.178}                                                  \\ 
\cline{2-2}
                                                                            & $\sigma_s$: 66.9/67.9/65.9            &                                                                    \\
\hline
\end{tabular}
\end{table}

\textbf{Backbone with the stride of 2.} Our backbone modified the stride of 4 (S4) in DLA-34 to 2 (S2). This modification can retain more spatial information beneficial to small and crowded object detection. From the 1st and 4th entries of Table \ref{tab:abl}, we can observe that our modification improves the F1-measure by 3.5\% and 4.6\% under threshold $\sigma_{l}$ and $\sigma_{s}$, respectively. Moreover, the inference time of S4 and S2 is 25ms and 29ms per image (with the 1600-pixel longer side), respectively, implying that such a modification only takes a little inference time increase (4ms).

\textbf{CHN and LHN.} We separately test the effectiveness of the counting head and localization head networks. With the CHN only (see the 2nd entry), the crowd counting errors increase significantly compared with the model with both heads (see the 4th entry), which illustrates that the localization head provides complementary features for the counting task, reducing the errors (e.g., the MAE drops from 81.8 to 64.4). Besides, the localization performance of the LHN only (see the 3rd entry) is competitive with that of both heads, which shows that the counting head affects the localization task marginally. By comparing the 2nd, 3rd, and 4th entries, we conclude that integrating the CHN and LHN into an end-to-end dense prediction network and simultaneously performing the two tasks not only retains the localization performance but also improves the counting ability.

\section{Conclusion}
\label{sec:conclusion}
This paper has presented a dense point prediction method for crowd counting and localization. With the simple yet effective modified DLA, counting and localization head networks, and the corresponding loss functions, the model achieved outstanding performance compared with state-of-the-art methods. We believe our method can be a baseline for further research. One valuable future research is to study an image-aware or object-aware threshold to select the predicted keypoints rather than the simple thresholding process, which will benefit crowd localization performance.

\bibliographystyle{IEEEbib}
\bibliography{icme2021template}

\end{document}